%% file: acl_latex.tex
\pdfoutput=1

\documentclass[11pt]{article}

\usepackage[preprint]{acl}

\usepackage{times}
\usepackage{latexsym}

\usepackage[T1]{fontenc}

\usepackage[utf8]{inputenc}

\usepackage{microtype}

\usepackage{inconsolata}

\usepackage{graphicx}
\usepackage{booktabs}

\usepackage{amsmath}
\usepackage{amssymb}
\usepackage{mathtools}
\usepackage{amsthm}

\usepackage{bbm}
\usepackage{multirow}
\usepackage{graphicx}
\usepackage{amsmath}
\usepackage{amsthm}
\usepackage{booktabs}
\usepackage{algorithmic}
\usepackage{amsthm}
\usepackage{mathtools}
\usepackage{multirow}
\usepackage{amsmath}
\usepackage{tabularx}
\usepackage{booktabs}
\usepackage{enumitem}
\usepackage{bbm}
\usepackage{xcolor}
\usepackage[ruled,linesnumbered,vlined]{algorithm2e}
\usepackage{color, colortbl}

\input{math_commands.tex}

\definecolor{Gray}{gray}{0.95}

\title{LLM-Empowered Class Imbalanced Graph Prompt Learning\\ 
for Online Drug Trafficking Detection}

\author{
    Tianyi Ma\textsuperscript{1}, Yiyue Qian\textsuperscript{2}$^*$, Zehong Wang\textsuperscript{1}, Zheyuan Zhang\textsuperscript{1},\\
        \textbf{Chuxu Zhang\textsuperscript{3}, Yanfang Ye\textsuperscript{1}}\\
        \textsuperscript{1} University of Notre Dame, Indiana, USA\\
        \textsuperscript{2} Amazon, Washington, USA\\
        \textsuperscript{3} University of Connecticut, Connecticut, USA\\
        tma2@nd.edu, yyqian5@gmail.com, \{zwang43, zzhang42\}@nd.edu,\\ chuxu.zhang@uconn.edu, yye7@nd.edu
        }

\begin{document}

\maketitle
\footnotetext{The work is not related to the position at Amazon.} 
\begin{abstract}
As the market for illicit drugs remains extremely profitable, major online platforms have become direct-to-consumer intermediaries for illicit drug trafficking participants. 
These online activities raise significant social concerns that require immediate actions. 
Existing approaches to combating this challenge are generally impractical, due to the imbalance of classes and scarcity of labeled samples in real-world applications.
To this end, we propose a novel \textbf{L}arge \textbf{L}anguage \textbf{M}odel-empowered \textbf{Het}erogeneous \textbf{G}raph Prompt Learning framework for illicit \textbf{D}rug \textbf{T}rafficking detection, called \textbf{LLM-HetGDT}, that leverages LLM to facilitate heterogeneous graph neural networks (HGNNs) to effectively identify drug trafficking activities in the class-imbalanced scenarios.
Specifically, we first pre-train HGNN over a contrastive pretext task to capture the inherent node and structure information over the unlabeled drug trafficking heterogeneous graph (HG). 
Afterward, we employ LLM to augment the HG by generating high-quality synthetic user nodes in minority classes. 
Then, we fine-tune the soft prompts on the augmented HG to capture the important information in the minority classes for the downstream drug trafficking detection task.
To comprehensively study online illicit drug trafficking activities, we collect a new HG dataset over Twitter, called Twitter-HetDrug. 
Extensive experiments on this dataset demonstrate the effectiveness, efficiency, and applicability of LLM-HetGDT. 
Our code is available at \href{https://anonymous.4open.science/r/LLM-HetGDT-DB76/README.md}{https://github/LLM-HetGDT}. 
\end{abstract}

\section{Introduction}
\begin{figure}[t]
    \centering
    \includegraphics[width=0.95\linewidth]{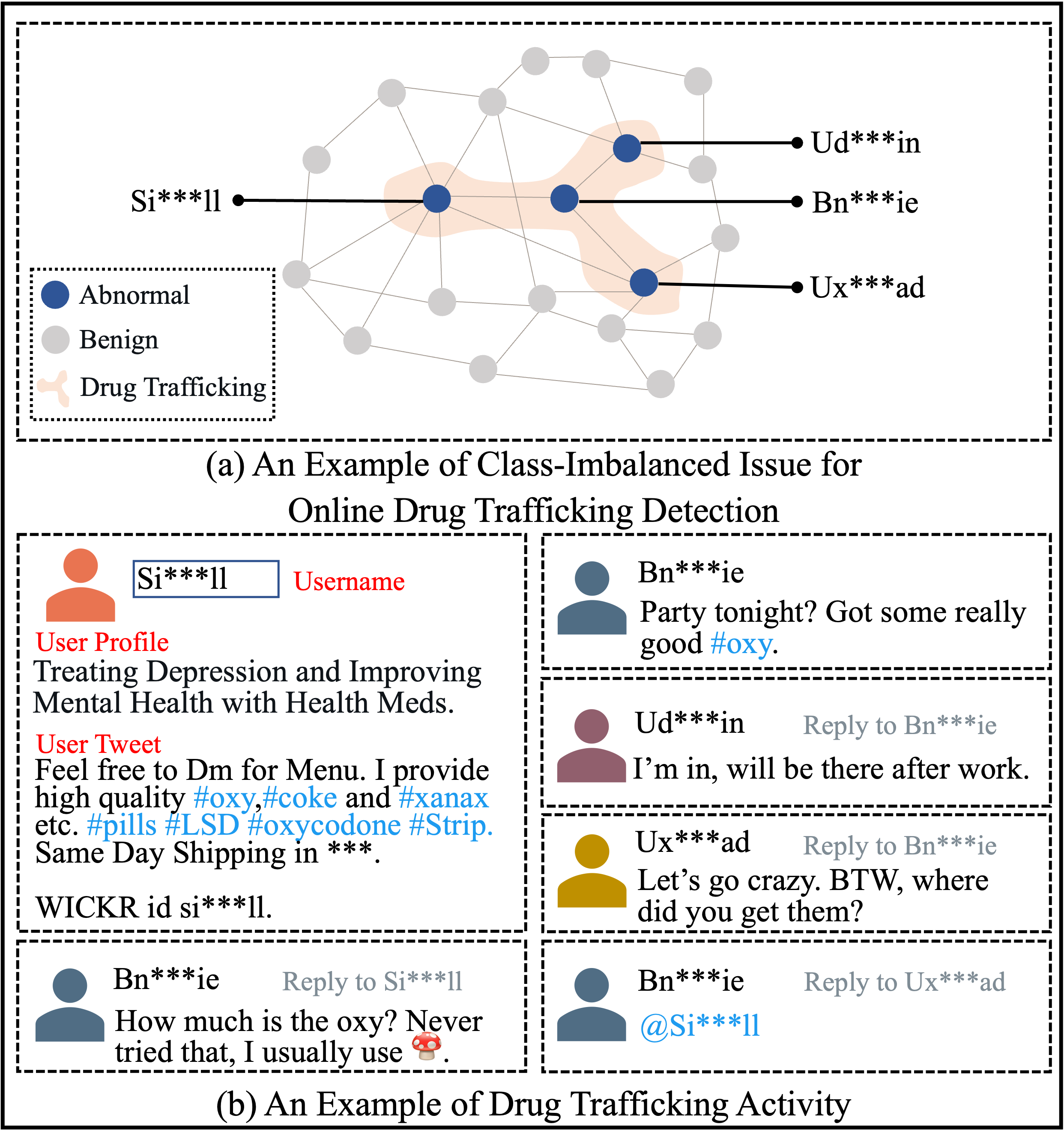}
    \vspace{-4mm}
    \caption{Illustration of drug trafficking activities among users on Twitter.}
    \label{fig: intro}
    \vspace{-5mm}
\end{figure}

As the market for illicit drugs (e.g., methamphetamine and heroin) remains immensely profitable, the crime of drug trafficking has continually evolved with modern technologies, e.g., online platforms. 
Recent studies~\cite{qian2021distilling, HyGCL-DC, zhang2020dstyle} have demonstrated that the major online platforms, e.g., Twitter/X, Facebook, and Instagram, have become direct-to-consumer intermediaries for illicit drug trafficking. 
As depicted in Figure~\ref{fig: intro} (b), illicit drug trafficking participants can effortlessly create accounts on these platforms to promote their products through drug-related posts and quickly attract other abnormal users to share or trade drugs on social media platforms.
Consequently, these drug-related behaviors among users naturally form drug trafficking networks on social media platforms, and these illegal drug-related activities have turned into a societal concern due to their catastrophic consequences, from violent crimes to public health (e.g., over 107,941 drug overdose deaths occurred in 2022 in the U.S.~\cite{centersdrug}).
To combat illicit drug trafficking online, there is an imminent need to develop novel and effective techniques for detecting drug trafficking activities on social media. 

To address this challenge, existing studies can be categorized into two approaches: (i) large language models (LLMs)-based frameworks~\cite{hu2024knowledge, hu2023unveiling} and (ii) graph-based methods ~\cite{li2019machine, mackey2018solution, HyGCL-DC}.
However, these two approaches still face limitations, such as class-imbalanced and label-scarcity issues.
On one hand, LLM-based methods may struggle to efficiently identify drug trafficking activities on social networks because of the latency limitation.
This inefficiency is exacerbated due to the class-imbalanced issue. 
Specifically, as illustrated in Figure~\ref{fig: intro} (a), merely four out of twenty users are involved in drug trafficking activities, and LLM-based methods require a significant amount of time to identify drug trafficking participants among huge social networks, making it impractical in real-world applications. 
Moreover, LLM-based methods primarily utilize user-level information, such as user profiles and posts for drug trafficking detection, while ignoring the rich and complex relationship information, e.g., users following others and replying to posts, etc.
For example, \citet{hu2024knowledge} utilizes anchor posts to identify drug-related users on Instagram, while ignoring contextual relationships between users.
As shown in Figure~\ref{fig: intro} (b), it is challenging to classify the users "Ud***in" and "Ux***ad" by analyzing their posts. 
However, by examining the rich contextual relationships between users, such as their replies to the known drug buyer "Bn***id", we can successfully identify them as drug users.
On the other hand, graph-based methods typically study drug trafficking in balanced data settings, which {fail to consider inherent class-imbalanced issues}, making it difficult for real-world scenarios. 
In addition, they require sufficient labeled samples for training, which is time and labor-intensive, and ignore the rich information in the unlabeled data.
This naturally raises the following question: \textit{How can we design an effective and efficient framework to detect drug-related users on social networks?}

To this end, we design a novel \textbf{L}arge \textbf{L}anguage \textbf{M}odel-enhanced \textbf{Het}erogeneous \textbf{G}raph Prompt Learning framework called \textbf{LLM-HetGDT} for \textbf{D}rug \textbf{T}rafficking detection.
Specifically, we leverage heterogeneous graphs (HGs) to model the complex relationships among entities, i.e., users, posts, and keywords, and further introduce the "pre-train, prompt" paradigm over heterogeneous graph neural networks (HGNNs) to comprehensively study drug trafficking activities on social media.
In the pre-train stage, we train an HGNN for the contrastive pretext task to capture the rich structure information in unlabeled data. 
Afterward, inspired by the success of LLMs in natural language understanding and generating synthetic data~\cite{zhang2024diet, tang2023does, long2024llms, zhang2024mopi, zhang2024ngqa}, we leverage LLMs to augment the original HG by generating synthetic user nodes in minority classes, i.e., drug trafficking participants, enabling the model to learn valuable information from drug trafficking participant nodes. 
We further fine-tune three types of prompts, i.e., node, structure, and drug trafficking prompts on the augmented HG to facilitate the pre-trained HGNNs to learn the node representations for downstream tasks. 
To conclude, our work makes the following contributions:
\begin{itemize}[leftmargin=*]
    \item \textbf{Novelty:} We develop a novel framework called LLM-HetGDT that integrates HGNNs with LLMs for heterogeneous graph representation learning to effectively address class imbalance and label scarcity issues in illicit drug trafficking detection.
        To the best of our knowledge, this is the \textbf{first work} that employs heterogeneous graph prompt learning and LLMs to study online drug trafficking activities with limited labeled samples under class-imbalanced settings. 
    \item \textbf{New Data:} To comprehensively study drug trafficking activities, we collect a new drug trafficking dataset from Twitter and further construct a heterogeneous graph called Twitter-HetDrug,  which contributes to research communities of drug trafficking and (class-imbalanced) graph learning. 
    \item \textbf{Effectiveness:} Comprehensive experiments on Twitter-HetDrug demonstrate the effectiveness of our proposed method in comparison with state-of-the-art baseline methods.
\end{itemize}

\section{Backgrounds}
\noindent\textbf{Preliminary.}
A heterogeneous graph is defined as a network $\gG = (\gV, \gE, \gX, \gA, \gR)$, where  $\gV$ is a node set with node type set $\gA$ and  $\gE$ is the edge set with relation type set $\gR$.  
Here, $\gX = \{\rmX^A | A\in\gA \}$ represents a set of the node attribute feature matrix $\rmX^A \in \mathbb{R}^{N_A\times d_A}$ with node size $N_A$ and dimension $d_A$ for node type $A\in\gA$.
The target node set $\gV^T$ is a set of nodes from node type $T\in \gA$ for node classification. 
The class imbalance ratio $\text{CIR}_{i,j} = |\gV_{C_i}^T|/|\gV_{C_j}^T|$, where $\gV_{C_i}^T$ and  $\gV_{C_j}^T$ denote the set of nodes belongs to class $C_i$ and $C_j$, respectively. If $\text{CIR}_{i,j}>1$, then $C_i$ is called the majority class, and $C_j$ is called the minority class, vice versa. 

\noindent\textbf{Related works.} 
We provide relevant works to tackle the class imbalance issues in this section. 
Existing works against the class imbalanced issue in graphs can be roughly divided into three categories, i.e., post-hoc correction~\cite{kang2019decoupling, tian2020posterior}, loss modification~\cite{mohammed2020machine, kang2019few}, and re-sampling~\cite{chawla2002smote}. 
For example, GraphSmote~\cite{zhao2021graphsmote} and ImGAGN~\cite{qu2021imgagn} introduce SMOTE~\cite{chawla2002smote, qian2025adaptive} and GAN~\cite{goodfellow2020generative} in graph augmentation, respectively, to generate synthetic minority nodes while learning the topological structure distribution at the same time~\cite{gao2023semantic}.  
DR-GCN~\cite{shi2020multi} and RA-GCN~\cite{ghorbani2022ra} propose re-weighting mechanisms to balance the loss of minority and majority classes. 
Motivated by existing re-sampling techniques, we propose an oversampling framework in the prompt tuning stage to obtain more effective prompts for drug trafficking detection tasks. 
The previous methods primarily over-sample synthetic nodes from the attribute feature level, potentially introducing redundant or harmful information.
In light of this, this work employs large language models to generate synthetic nodes from the text attribute level to obtain more accurate and valuable synthetic nodes.
Additional related work for graph-based methods is provided in Appendix~\ref{appendix: graph}.

\noindent\textbf{Problem Definition.} \textit{ \textbf{Class-imbalanced Graph Prompt Learning for Drug Trafficking Detection.}} Given a class-imbalanced heterogenous graph $\gG = (\gV, \gX, \gE, \gA, \gR)$ built on real-world drug trafficking data over social media, the objective is to build a graph prompt learning model to detect drug trafficking participants. 
\vspace{-1mm}
\section{Methodology}
\begin{figure*}[t]
   \centering
    \includegraphics[width=0.96\linewidth]{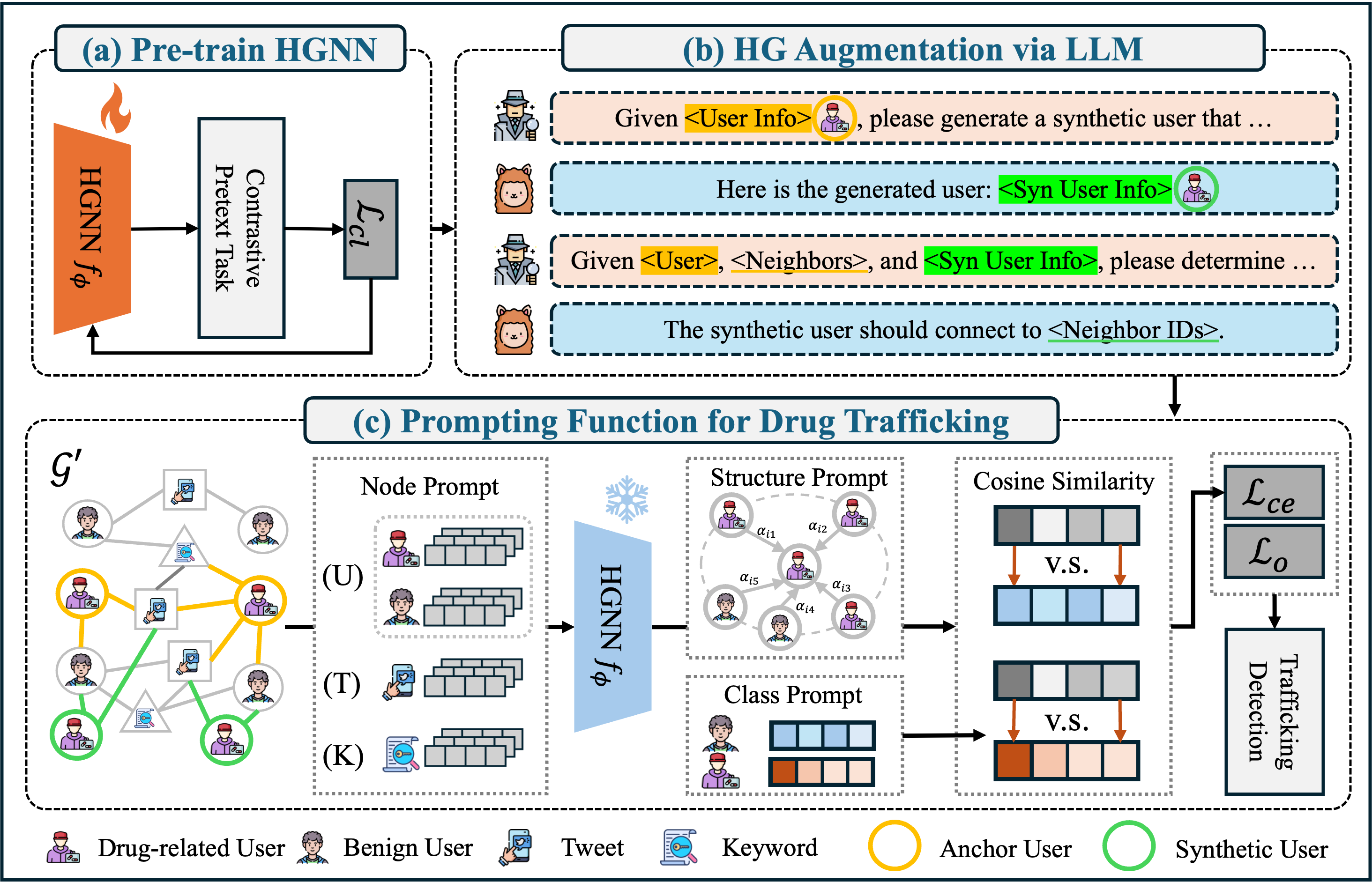}
    \vspace{-3mm}
    \caption{The overall framework of LLM-HetGDT:  (a) It pre-trains HGNN with the contrastive pretext task. (b) It leverages LLM to generate synthetic users and connections between synthetic users and neighbors of the original users, forming an augmented HG $\gG'$; (c) LLM-HetGDT injects node prompt to node attribute features and feeds the augmented HG $\gG'$ into the pre-trained HGNN to obtain the target node embeddings. Afterward, it augments the target node embeddings with structure prompt and further computes the similarity between node embeddings and class prompt to obtain the classification loss for optimization.
    }\label{fig: framework}
    \vspace{-4mm}
\end{figure*}

In this section, we present the details of LLM-HetGDT, which includes two key steps: 
(i) pre-train HGNN,
and (ii) LLMs-enhanced prompt tuning for drug trafficking detection. 
Mention that the original HG contains text-attributed features, and we leverage SentenceBert~\cite{reimers2019sentence} to encode text attributes $\gT = \{t_i | v_i\in \gV \}$ into attribute features $\gX = \{x_i| v_i \in \gV \}$. 
The overall framework is illustrated in Fig.~\ref{fig: framework}.

\subsection{Pre-train HGNN}
As we mentioned above, previous studies face the label-scarcity issue, and underestimate the rich information in the unlabeled data.
Therefore, inspired by the success of "pre-train, prompt" in natural language processing~\cite{liu2023pre}, we propose a novel LLM-enhanced HG prompt learning framework that pre-trains HGNN via contrastive pretext tasks to capture rich contextual and structure information in HG, and further fine-tune prompts to reformulate the downstream detection task better align with the pretext task.
\\\noindent\textbf{Pre-training Stage.}
Our framework is applicable to any HGNN. 
In this work, we utilize a cross-view contrastive learning method~\cite{wang2021self}, as the example to map target nodes $\gV^T$ into low-dimensional representations $\rmZ\in\sR^{N_T\times d}$, where $N_T$ is the size of target nodes and $d$ is the latent dimension.
Given an HG $\gG=(\gV, \gE, \gX, \gA, \gR)$ that models drug trafficking activities on social media, we first adopt a type-specific mapping layer to project all the node attribute features $\gX$ into the same space: $\tilde\rmX^A = \rmX^A W^A + b^A$, where $\rmX^A \in\sR^{N_A\times d_A}$,  $W^A\in\sR^{d_A \times d}$, and $b^A\in\sR^{d}$ denote the attribute feature matrix, projection matrix, and the bias vector for node type $A\in\gA$, respectively.
Afterward, we employ meta-path view and network schema view encoders~\cite{wang2021self} to map each target node $v_i\in\gV^T$ into two embeddings $z_i^{mp}\in\sR^d$ and $z_i^{sc}\in\sR^d$, respectively.
These two embeddings are then fed into the pretext objective function to optimize the HGNN. 

\noindent\textbf{Pre-train Optimization.} 
Inspired by GCL~\cite{you2020graph, liu2022revisiting}, we leverage the temperature-scaled contrastive loss as the objective function for the pretext task. 
The objective function is formulated as follows:
\begin{equation}
    \gL_{cl} = - \frac{1}{N} \sum_{v_i\in\gV_T} \frac{\exp(\delta_{i,i}/\tau)}{\sum_{j\ne i} \exp(\delta_{i,j}/\tau) + \exp(\delta_{i,i} / \tau)}, \label{eq: nt-xent}
\end{equation}
where $\delta_{i,j}$ is the cosine similarity between $z_i^{mp}$ and $z_j^{sc}$,
$\tau$ is the temperature hyper-parameters. 
We adopt $z^{mp}$ as the target node representations $\rmZ$ since the target type nodes explicitly participate in aggregating $z^{mp}$.
With the pre-trained HGNN, we further devise a LLMs-enhanced prompt tuning framework for drug trafficking detection tasks. 
\vspace{-2mm}
\subsection{Prompt Tuning for Drug Trafficking Detection}
\subsubsection{HG Augmentation via LLM}\label{sec: hg augmentation}
As drug trafficking participants are inactive and rare on social media, the HG to model real-world online drug trafficking activities is a class-imbalanced network.
This class imbalance issue may lead to a performance bias, as the majority class tends to outperform the minority class due to their more significant number of labeled nodes~\cite{zhao2021graphsmote, shi2020multi, liu2023survey}.
To tackle this issue, existing works~\cite{chawla2002smote,  park2022graphens, zhao2021graphsmote} mainly balance the data distribution through over-sampling techniques from the attribute feature level.
However, due to the limited number of nodes in the minority class and the inevitable information loss in feature extraction from raw textual information to attribute features~\cite{li2020sentence}, the sampled synthetic nodes may introduce redundant or even harmful information for downstream tasks. \\
\noindent\textbf{Synthetic User Generation.} 
To handle the above challenge, inspired by the success of LLMs in generating synthetic data~\cite{tang2023does, long2024llms},
we leverage an LLM to generate synthetic minority users in natural languages, thereby balancing the data distribution toward minority class (drug trafficking participants).
Specifically, for each labeled user node $v_i$ in the minority class, we feed the user text content $t_i$ into LLM to obtain a synthetic node $v_i'$ with corresponding user text content $t_i'$. 
Formally, we have:
\begin{align}
    t_i' = \textbf{LLM}(\textbf{Prompt}_\text{node}(t_i)),
\end{align}
where user text content $t_i$ contains the user $i$'s information, including username, user ID, and user profile. 
Afterward, we employ SentenceBert over user text content $t_i'$ to obtain the attribute feature $x_i'$ for synthetic node $v_i'$:
\begin{equation}
    x_i' = \textbf{SentenceBert}(t_i').
\end{equation}
The prompt template for synthetic user generation is listed in Appendix Table~\ref{tab: node prompt}.
\\
\noindent\textbf{Edge Generation Among Synthetic Nodes.} 
Up to now, we have obtained a set of synthetic nodes corresponding to the labeled user nodes in the minority class. 
However, these nodes are still isolated in the HG $\gG$.
Intuitively, for each pair of labeled user node $v_i$ and its corresponding synthetic node $v_i'$, the synthetic node $v_i'$ should have connections with the neighbors of the labeled node $v_i$.  
However, connecting the synthetic node $v_i'$ with every neighbor of node $v_i$ is not feasible, as the synthetic node contains different information than the original node.  
Therefore, to select more appropriate and informative neighbors for synthetic nodes, we further employ LLM to generate edges between synthetic node $v_i$ and the neighbors of the original node $v_i$, i.e., $\gN(v_i)$. 
Mathematically,  we have:
\begin{align}
    e = \textbf{LLM}(\textbf{Prompt}_\text{edge}( t_i', t_j)).
\end{align}
The prompt template for edge generation is listed in Appendix Table~\ref{tab: edge prompt}.
Through aforementioned steps, we have an augmented HG $\gG' = \{\gV', \gE', \gX', \gA, \gR \}$,
where $\gV' = \gV \cup \{v_i' | v_i\in\gV_c\}, 
    \gE' = \gE \cup \{ (v_i', v_j) | v_j\in \gN^A(v_i'), A\in\gA\}, \text{and }
    \gX' = \gX \cup \{x_i' | v_i\in\gV_C\}$.
Here, $\gV_C$ is a set of nodes in minority class $C\in\gC$, $\gN^A(v_i)$ denotes the neighbors of node $v_i$ in node type $A$. 
Notice that we merely sample synthetic nodes for labeled user nodes in the minority class. 
Moreover, in the pre-train stage, as the label information is unavailable, we only employ the original HG $\gG$ for pre-train. 
The augmented HG $\gG'$ is exclusively used to tune prompts to learn valuable information among drug trafficking participants. 
\subsubsection{HG Prompting Function Design}
As the training objective gap between the pretext contrastive task and downstream task, the pre-trained HGNN may fail to identify the drug trafficking participants effectively. 
Hence, we introduce a novel prompting function, integrated with node, structure, and drug trafficking prompts, to bridge the training objective gap between the pretext task and the downstream drug trafficking detection task.

\noindent\textbf{Node Prompt.} 
Previous studies~\cite{fang2024universal, ma2024hetgpt, sun2023all} have demonstrated that incorporating a small number of trainable parameters directly into the node feature space can effectively tailor pre-trained node representations, facilitating the knowledge transfer from pretext task to downstream task. 
However, existing methods primarily introduce feature weight-based prompt~\cite{liu2023graphprompt, yu2024hgprompt} or learnable independent basis tokens~\cite{sun2022gppt, ma2024hetgpt} as prompt, which majority classes samples can simply dominate the prompt tuning process. 
To tackle this issue, besides the LLM-enhanced over-sampling technique in Section~\ref{sec: hg augmentation}, we devise a set of independent basic tokens for each class, served as class-specific augmentations to the original target node attribute features. 
Specifically, for each class $C\in\gC$, the class-specific feature tokens $\rmF^C\in\sR^{K\times d_T}$ are defined as:
$\rmF^C = [\rvf_1^C, \rvf_2^C, ..., \rvf_K^C]^\top $,
where $\rvf_*^C\in\sR^{d_T}$, $d_T$ is the attribute feature dimension for target node type $T$ and $K$ is a hyper-parameter.
Afterward, the feature tokens $\rmF^C$ are injected into the target node attribute features $\rmX$:
\begin{align}
    \bar\rmF^C &= \textbf{Att}(\rmX, \rmF^C), \\
     \hat\rmX_{i, :} &= \rmX_{i, :} +  \textbf{PMA}(\{\bar\rmF^C| C\in\gC\}), \label{eq: prompt node}
\end{align}
where $\bar\rmF^C\in\sR^{d_T}$ is the linear combination of feature tokens $\rmF^C$ through an attention mechanism $\textbf{Att}(\cdot)$ and $\textbf{PMA}(\cdot)$ is the set transformer function~\cite{lee2019set}.
As attention and set transformer mechanisms are not the main contribution of this paper, we refer readers to Appendix~\ref{appendix: attention} and~\ref{appendix: set transformer} for more detail about these two mechanisms.
The key insight behind the class-specific feature tokens is to separate the prompt for each class to alleviate the influence of class imbalance and further capture the unique characteristics of each class equally. 
Besides, considering the heterogeneity of HGs, we also employ a set of tokens for the rest of node types $A\in\gA /T$ :
$\rmF^A = [\rvf_1^A, \rvf_2^A, ..., \rvf_K^A]^\top,$
where $\rvf_k^A\in\sR^{d_A}$, $d_A$ is the attribute feature dimension for node type $A$.
Then, we reuse the attention mechanism $\textbf{Att}(\cdot)$, and inject the prompt into the attribute feature $\rmX^A$:
\begin{equation}
    \hat\rmX_{i,:}^A =  \rmX_{i,:}^A + \textbf{Att}(\rmX_{i,:}^A, \rmF^A).
\end{equation}
Subsequently, the prompted node features $\hat\gX = \{\hat\rmX^A | A\in\gA\}$, and the augmented HG $\gG'$ are fed into pre-trained HGNN to obtain the target node embeddings $\rmZ\in\sR^{N_T\times d}$. \\
\noindent\textbf{Structure Prompt.} 
Merely relying on the feature prompt may not be sufficient to transfer the learned relation information from pre-training to drug trafficking detection tasks. 
Prior studies~\cite{HyGCL-DC, qian2021distilling} have demonstrated that drug trafficking participants are more likely to connect with other participants on social media.
Therefore, we propose a structure prompt to leverage the expressive meta-path based neighborhood information for robust classification performance.   
Particularly, for each target node $v_i$, we employ the set transformer function $\textbf{PMA}(\cdot)$ to aggregate the neighbor node embeddings with respect to meta-path $P\in\gP$:
\begin{align}
     \rmH^P_{i, :} = \text{PMA}(\{\rmZ_{j, :} | v_j\in\gN^P(v_i)\}). \label{eq: structure}
\end{align}
Here a meta-path $P$ is a path defined in the form of $A_1 \xrightarrow{R_1} A_2 \xrightarrow{R_2} \dots \xrightarrow{R_l} A_{l+1}$, where $A_i\in\gA$ and $R_i\in\gR$. It describes a composite relation $R = R_1 \circ R_2 \circ \dots \circ R_l$ between nodes $A_1$ and $A_{l+1}$, where $\circ$ denotes the composition operator.
Afterward, we employ semantic attention~\cite{wang2019heterogeneous}, i.e., $\textbf{SM-Att}(\cdot)$, to aggregate the structure embeddings as structure tokens $\rmS$ among meta-paths $\gP$:
\begin{align}
    \rmS_{i, :} = \textbf{SM-Att}(\{\rmH_{i, :}^P | P\in\gP\}).
    \vspace{-3mm}
\end{align}
Again, we leave the detailed formulation of meta-path $\gP$ and semantic attention to Appendix~\ref{appendix: meta path} and~\ref{appendix: semantic attention}, respectively.

\noindent\textbf{Drug Trafficking Prompt.} Instead of the discrete class labels for drug trafficking participants and benign, we introduce the drug trafficking prompt, which contains a set of class prototype tokens, i.e., $\rmC = [\rvp, \rvn]$,
where  $\rmC\in\sR^{2\times d}$, and $d$ is the same dimension as the latent node embeddings. 
Here, $\rvp$ and $\rvn$ denote the prototype tokens for drug trafficking participants and benign, respectively. 
Moreover, each prototype token is initialized by the mean of labeled nodes embeddings:
\begin{equation}
    \rvp = \frac{1}{|\gV_\text{p}|} \sum_{v_i\in\gV_\text{p}} \rmZ_{i,:},  \rvn = \frac{1}{|\gV_\text{n}|} \sum_{v_i\in\gV_\text{n}} \rmZ_{i,:}.\label{eq: class prompt init}
\end{equation}
These task tokens serve as class prototypes for node classification, conveying the discrete labels to class-specific semantics~\cite{sun2023all, sun2022gppt, ma2024hetgpt}. 
The key idea behind the task prompt is shifting the node classification to pairwise similarity calculation between the node embeddings and prototype tokens, which can seamlessly align with the pretext contrastive task. 
Moreover, the prompt tuning process facilitates the prototype tokens to represent the unique characteristics of each class, even some have fewer labeled samples. 

\subsubsection{Prompt Tuning}
With the drug trafficking prompts $\rmC$, structure prompt $\rmS$, and target node embeddings $\rmZ$, we employ an MLP to project them into sample embedding space.
Moreover, the structure prompt is injected into the target node embeddings to obtain the final node embeddings: 
$\rmZ' = \rmZ + \delta\rmS,$
where $\delta$ is a hyper-parameter.
To align with the pre-training stage, we adopt the same temperature-scaled contrastive loss in Equation~\ref{eq: nt-xent} as the objective function for node classification loss.
Moreover, we employ orthogonal constraints to keep the drug trafficking prompt orthogonal during prompt tuning:
\begin{equation}
    \gL_{o} = || \rmC \rmC^\top - \rmI||_F^2,
\end{equation}
where $\rmI$ is the identity matrix. Then, the final objective for prompt tuning is formulated as:
\begin{equation}
    \gL_{pt} = \gL_{ce} + \lambda\gL_{o}, \label{eq: pt loss} 
\end{equation}
where $\lambda$ is the trade-off hyper-parameters.
The pseudocode for the training procedure is provided in Appendix Alg.~\ref{alg: LLM-HetGDT}.

\section{Experiments}
\subsection{New Dataset for Drug Trafficking Detect}
To comprehensively study online drug trafficking activities in a real-world scenario, we build a new real-world HG dataset on Twitter, called \textbf{Twitter-HetDrug}, to analyze drug trafficking activities under a class-imbalanced setting, i.e., merely 11.9\% of users are drug trafficking participants.
Our data is collected through the official Twitter API from Dec. 2020 to Aug. 2021. 
More detailed discussion about the dataset (i.e., data collection, construction and annotation) and data statistics are provided in Appendix~\ref{appendix: data description} and Appendix Table~\ref{tab: data static}, respectively.

\begin{table*}[htbp]
    \setlength\tabcolsep{7pt}
    \renewcommand{\arraystretch}{0.7}
    \begin{tabular}{c|l|c|c|c|c|c|c}
    \toprule
      \multicolumn{2}{c|}{Setting} & \multicolumn{2}{c|}{Twitter-HetDrug 10\%} &  \multicolumn{2}{c}{Twitter-HetDrug 20\%} & \multicolumn{2}{|c}{Twitter-HetDrug 40\%} \\
      \midrule
      Group  & Model &  Macro-F1 & GMean &  Macro-F1 & GMean &  Macro-F1 & GMean  \\
       \midrule\midrule
      {G1} & GCN  & 55.74 & 53.92 & 58.38  & 56.43  & 60.94 & 59.72 \\

      {G2}   & HGT  & 60.95 & 59.88 & 63.85 & 63.70  & 67.08 & 66.37 \\
         {G3}  & HeCo   & 62.93 & 60.85 & 64.83  & 63.21 & 67.94  & 66.52 \\  
          {G4} & Over-sampling &  63.79  & 62.27 & 65.10 &  63.08 & 68.35 & 67.39 \\
     
         {G5}
         
          & PC-GNN  & 65.98  & 63.84 & 67.52  & 64.71  &  \underline{70.59}  & \underline{71.94} \\
          {G6} 
         & HGPrompt & \underline{66.85} & \underline{64.37} & \underline{67.97} & \underline{66.08}  & 69.72 & 69.42  \\
         {G7} & KG-GPT & 65.74 & 63.92 & 65.36 & 63.98 & 65.18 & 63.65 \\ 
         \midrule
\rowcolor{Gray}    Ours & LLM-HetGDT  & \textbf{70.52}  & \textbf{71.69} & \textbf{71.36} & \textbf{73.87} &  \textbf{74.06} &  \textbf{76.24}\\
           \bottomrule
    \end{tabular}
    \vspace{-3mm}
        \centering\caption{Performance comparison among various methods on Twitter-HetDrug for drug trafficking detection. Bolded numbers indicate the best performance, and underlined numbers represent runner-up performance.} \label{tab: main performance}
    \vspace{-4mm}
\end{table*}

\subsection{Experimental Setup}
\textbf{Baseline Methods.} To evaluate LLM-HetGDT, we compare it with Twenty-one state-of-the-art (SOTA) baseline methods, which are divided into seven groups: graph representation learning methods (G1), heterogeneous graph representation learning models (G2), heterogeneous graph self-supervised learning methods (G3), generic methods against class imbalance data (G4), graph models against the class imbalance issue (G5), graph prompt learning models (G6), and LLM-based methods (G7). 
A detailed discussion about each baseline methods is provided in Appendix~\ref{appendix: baseline}.
Due to the limited space, we only list the best model performance of each group in Table~\ref{tab: main performance}, and the detailed performance with standard deviation of each baseline method can be found in Appendix~\ref{appendix: complete experiments}.

\noindent\textbf{Experimental Settings.} 
All experiments are conducted under the environment of the Ubuntu 22.04 OS, plus Intel i9-12900K CPU, four Nvidia A40 Graphic Cards, and 48 GB of RAM. We train all methods with a fixed number of epochs (i.e., 500). 
We use Llama3.1 70B~\cite{llama3.1} as our LLM backbone. 
Besides, all supervised methods, i.e., in G1-G6, are trained five times, and the average performance on testing data is reported. 
We utilize Adam~\cite{kingma2014adam} as the optimizer and run a grid search on hyper-parameters for each model. 
Then, we report the best performance among the optimal hyper-parameters for each method.
Detailed discussion for experimental settings is provided in Appendix~\ref{appendix: experimental settings}.
For LLMs-based methods, we set the temperature to 0 to ensure the deterministic output.
Following existing works~\cite{liu2021pick, park2022graphens, goodfellow2020generative} in evaluating class imbalance classification, we adopt two metrics, i.e., Macro-F1 score (\textbf{Macro-F1}) and GMean score (\textbf{GMean}), to evaluate all models. 
\subsection{Performance Comparison}
Table~\ref{tab: main performance} lists the performance of selected baseline methods and LLM-HetGDT over Twitter-HetDrug with different percentages, i.e., 10\%, 20\%, and 40\%, of training data for drug trafficking detection. 
Besides, we use 10\% samples for validation and the rest for testing.
According to Table~\ref{tab: main performance}, we can conclude that: 
(i) HG methods outperform graph methods, i.e., all heterogeneous graph representation learning models in G2 and G3 show better performance than graph-based methods in G1. 
This result indicates \textbf{the necessity of heterogeneous graphs for drug trafficking detection tasks.}
(ii) Class-imbalanced methods in G4 and G5 outperform all (heterogeneous) graph representation learning methods in G1-G3, demonstrating that \textbf{class imbalance is a significant challenge for online drug trafficking detection tasks}, where traditional semi-supervised (heterogeneous) graph representation learning may struggle to learn the patterns from minority classes.
(iii) HGPrompt is the best baseline model in low-labeled settings, i.e., 10\% and 20\%, however in high-labeled settings, the best baseline model is PC-GNN. 
This indicates the superiority of graph prompt learning in low-labeled settings, while methods that are specifically designed for class-imbalanced issues in G4 and G5 may fail to model the minority classes due to the limited labeled data.  
However, in high-labeled settings, methods specifically designed for class-imbalanced data may learn enough information from minority classes.
This validates \textbf{the need to design effective techniques for class imbalance issues in low-label settings.} 
(v) LLM-based methods, i.e., in G7, have consistent performance in different label settings, as we do not fine-tune LLMs, and merely prompt it for prediction. 
Despite this, LLM-based approaches still outperform some supervised methods in G1-G4, demonstrating the superiority of natural language understanding and generation ability of LLMs. 
This is one of the reasons we propose to leverage LLMs to generate synthetic user nodes to tackle class-imbalance issue in drug trafficking detection tasks.
(vi) Our model LLM-HetGDT gains the best performance by comparison with all baseline models, which demonstrates \textbf{the effectiveness of our model over Twitter-HetDrug.}
\subsection{Ablation Study}

\vspace{-1mm}
\subsubsection{Prompt Component Analysis}
\begin{figure}[t]
    \begin{center}
        \includegraphics[width=0.48\textwidth]{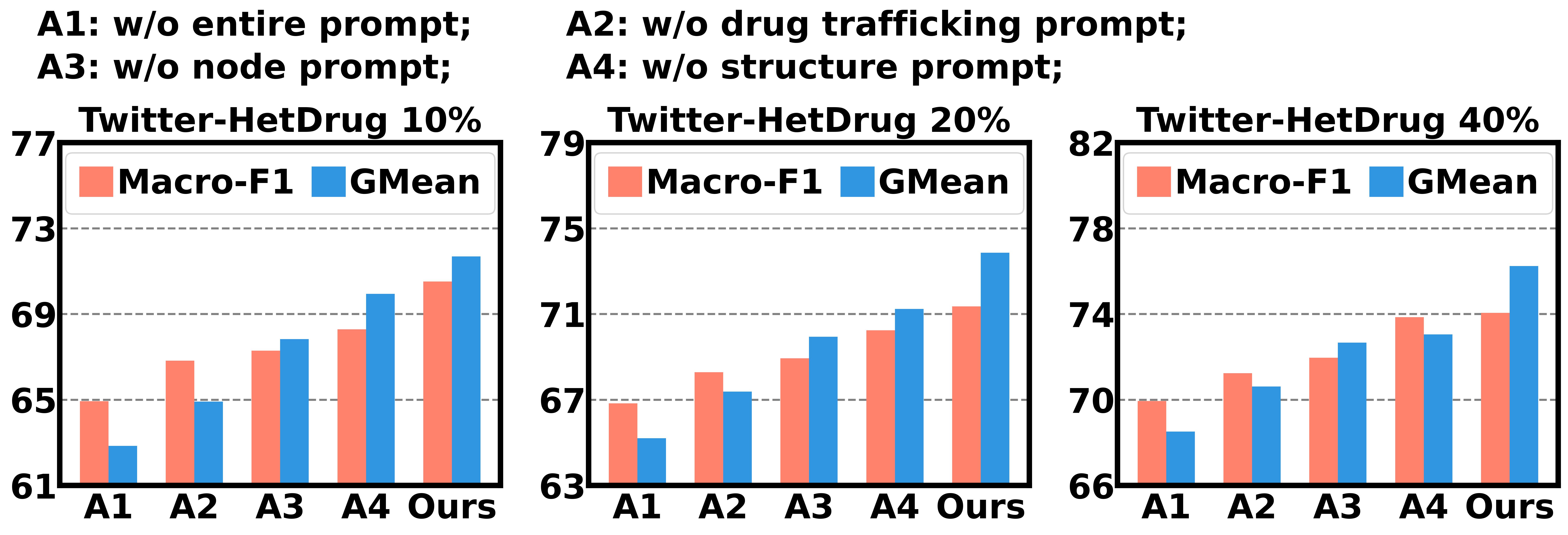}
    \end{center}
    \vspace{-4mm}
    \caption{Performances of model variants over Twitter-HetDrug with training splits 10\%, 20\%, and 40\%.}\label{fig: ablation}
    \vspace{-3mm}
\end{figure}

To show the effectiveness of each prompt in LLM-HetGDT, we further analyze the contribution of each component, i.e., entire prompts (A1), drug trafficking prompt (A2), node prompt (A3), and structure prompt (A4), by removing it separately, as illustrated in Fig~\ref{fig: ablation}.
First, we remove all three prompts, and the significant drop in performance indicates the effectiveness of our proposed framework. 
Afterward, we remove the drug trafficking prompt (A2), node prompt (A3), and structure prompt (A4), respectively. 
The performance of A2 decreases obviously, showing the necessity of the prototypes to capture the complex characteristics among drug trafficking participants for classification tasks.  
Moreover, the decline in the performance of A3 and A4 indicates the effectiveness of the node and structure prompts in our model.

\begin{table}[t]
    \centering
    \setlength\tabcolsep{4pt}
    \renewcommand{\arraystretch}{0.7}
    \begin{tabular}{c|c|c|c|c}
        \toprule 
         Class & \# nodes  & PC-GNN & HGPrompt & Ours \\
         \midrule
        Neg.  &  25,402  & \underline{88.39} & 87.08 &  \textbf{90.47}\\
        Pos. & 3,437  & 43.57 & \underline{46.62}& \textbf{50.65} \\
        \midrule
Overall & 28,839 & 65.98 & 66.85 & 70.52\\
        \bottomrule
    \end{tabular}
    \vspace{-3mm}
        \caption{Macro-F1 among models over Twitter-HetDrug-10\%. Pos. and Neg. indicates drug participants and benign users, respectively.}
    \label{tab: ablation minority}
     \vspace{-3mm}
\end{table}
\subsubsection{Minority Class Analysis}
To clearly show the improvement in the minority class, Table~\ref{tab: ablation minority} lists Macro-F1 among some models for Twitter-HetDrug-10\%. We find out that all models gain excellent performance over the majority class while they fail to achieve good performance over the minority class (participants). Our proposed method outperforms all baseline methods over both minority and majority classes, which again demonstrates the effectiveness of our model.
\subsection{Deployment Evaluation}
To validate the applicability of our model, we deploy LLM-HetGDT in a live manner to detect drug trafficking activities over Twitter in Jan. 2025, and compare with two SOTA drug trafficking detection models, i.e., MetaHG~\cite{qian2021distilling} and KG-GPT~\cite{hu2024knowledge}.
The detailed settings of the deployment are provided in Appendix~\ref{appendix: deployment}, and the results are shown in Table~\ref{tab: live evaluation}.
According to the table, we conclude that our proposed method outperforms the baseline methods in every aspect, which indicates the effectiveness, efficiency, and applicability of our model for online illicit drug trafficking detection tasks.
\begin{table}[htbp]
    \vspace{-2mm}
        \setlength\tabcolsep{3.5pt}
        \renewcommand{\arraystretch}{0.7}
        \begin{tabular}{c|c|c|c}
            \toprule
            Model & Macro-F1 $\uparrow$ & GMean $\uparrow$ & Avg. Time $\downarrow$ \\ 
            \midrule\midrule 
            MetaHG & 61.73 & 62.48 & \underline{0.39s} \\ 
            KG-GPT & \underline{64.24} & \underline{63.18} & 4.72s \\ 
            \midrule
           \rowcolor{Gray} Ours & \textbf{66.38} & \textbf{65.35} & \textbf{0.18s}  \\
            \bottomrule
        \end{tabular}
        \vspace{-3mm}
          \centering \caption{The performance of drug trafficking detection methods for deployment evaluation. } \label{tab: live evaluation}
        \vspace{-3mm}
    \end{table}

\subsection{Case Study}\label{sec: case study}
We analyze the synthetic users generated by LLM. 
Table~\ref{tab: case study syn user} illustrates an example of an input prompt with original user context and the output synthetic user generated by LLM.
According to Table~\ref{tab: case study syn user}, the original user profile contains three key phrases, i.e., "online dispensary," "premium lab-tested psychedelic products," and hashtags "\#LSD \#420 \#acid" (highlighted in green, red, and blue, respectively), that help to identify the user "yas***exb" as a drug trafficking participant. 
 In the output, the LLM effectively captures these key phrases that best depict the synthetic user and introduce new information, i.e., "Get ready to ..." and street name hashtags to ensure variability, which aligns with our prompt design. We conduct another case study for edge generation in Appendix~\ref{appendix: case study}.
\begin{table}[t]
    \centering 
    \renewcommand{\arraystretch}{0.6}
    \begin{tabular}{p{0.95\linewidth}}
        \toprule
        \textbf{INPUT: }\\
        \midrule
        \textbf{Instruction:}
        <The same instruction in Table~\ref{tab: node prompt}>. \\
        \textbf{Context:} \\
        Username: yas***exb; \\
        User ID: @dru***cy; \\
        User Profile: \textcolor{green}{Online dispensary}, \textcolor{red}{premium lab-tested psychedelic products,} must be 18+, same day shipping in ***. DM me for menu. \textcolor{blue}{\#LSD \#420 \#Meth \#acid.} \\
        \midrule
        \textbf{OUTPUT: }\\
        \midrule
        Username: Psy***ete; \\
        User ID: @ss***ete; \\
        User Profile: \textcolor{green}{Your one-stop shop} for \textcolor{red}{lab-tested psychedelic products.} Get ready to trip with the selection of \textcolor{red}{premium psychedelic products.}
       \textcolor{blue}{\#lsdtrip \#psychedelics \#acid \#meth \#420.}\\
       \bottomrule
    \end{tabular}
    \vspace{-3mm}
     \caption{Example of synthetic user generated by LLM.} \label{tab: case study syn user}  
    \vspace{-4mm}
\end{table}

\section{Conclusion}
In this study, to handle the class-imbalanced and label-scarcity issues in existing illicit drug trafficking detection methods, we propose a novel LLM-enhanced heterogeneous graph prompt learning framework, called LLM-HetGDT, to detect online drug trafficking participants in real-world scenarios. 
Specifically, we leverage LLMs to augment the heterogeneous graph (HG) by generating synthetic nodes in minority class, i.e., drug trafficking participants, facilitating the heterogeneous graph neural networks (HGNNs) to learn valuable information in minority class. 
To comprehensively study the drug trafficking activities on social media, we collect a new HG dataset over Twitter, called Twitter-HetDrug.
Extensive experiments on the dataset demonstrate the effectiveness, efficiency, and applicability of our framework. 

\section{Limitations}
In our work, we follow existing studies~\cite{HyGCL-DC, qian2021distilling} that utilize SentenceBert to encode the text attributes. 
We acknowledge that there are alternatives available in the literature that may have better performance. 
Moreover, we leverage open-source LLMs, i.e., Llama3.1 70B, to generate synthetic users and edges, which may have a lower capability compared to other closed-source LLMs, e.g., GPT-4o, DeepSeek-R1, etc.
Despite this, our experimental results are still highly desirable. 
It remains uncertain whether more powerful text encoders or LLMs can further improve the performance of our proposed method to some extent. 
We regard these two limitations as promising directions for future work.
\section{Ethical Considerations}
Our method serves as a powerful tool for online drug trafficking detection. 
However, there is potential for the generated content to include biased or harmful information. To mitigate this risk, more careful prompt design, including clear instructions and guidelines, can help steer the LLMs toward generating positive and accurate content. Additionally, users must be mindful of ethical concerns such as bias in the data and ensure responsible use of the tool in different applications.

\bibliography{citations}
\clearpage
\appendix
\textbf{\huge{Technical Appendix}} 
\section{Detailed Related Works}\label{appendix: graph}
\textbf{Heterogeneous Graph Neural Networks.} 
In the past few years, several graph neural networks (GNNs)~\cite{wang2024gft,ma2025adaptive, wang2025training, wang2025neuralgraphpatternmachine, qian2024dual, wang2024learning, wang2023heterogeneous} have been proposed to study representations on graphs.
Unlike GNNs, heterogeneous graph neural networks (HGNNs) aim to study node representations on heterogeneous graphs that contain different attribute feature types~\cite{wang2022survey,  fu2020magnn, bing2023heterogeneous, zhao2024dahgn}. 
For example, HAN~\cite{fu2020magnn} converts a heterogeneous graph into multiple metapath-based homogeneous graphs and leverages hierarchical attention to encode node-level and semantic-level structures. HetGNN~\cite{zhang2019heterogeneous} introduces random walk~\cite{perozzi2014deepwalk} to sample node neighbors and further fuses their features using LSTMs~\cite{hochreiter1997long}. 
HGT~\cite{hu2020heterogeneous} proposes a transformer-based architecture for web-scale heterogeneous graphs.
However, these models require sufficient high-quality labeled data for training~\cite{jiang2021pre, yoon2022zero}, which is labor-intensive and impractical for online drug trafficking detection. 
Unlike existing works, we propose a graph prompt learning framework to address the label scarcity issue for drug trafficking detection on social media. \\
\noindent\textbf{Prompt-based Learning on Graph.} Recently, a new paradigm known as "pre-train and prompt"~\cite{liu2023pre} has gradually gained attention in the natural language processing field. This paradigm reformulates downstream tasks to resemble pre-train tasks via prompts among input texts~\cite{gu2021ppt}. This not only bridges the gap between pretraining and downstream tasks but also instigates further research integrating prompting with pre-trained graph neural networks~\cite{sun2023all, ma2024hetgpt}. 
For example, GraphPrompt~\cite{liu2023graphprompt} utilizes the subgraph similarity and a learnable prompt for downstream few-shot learning. 
GPPT~\cite{sun2022gppt} introduces a template to align the pretext task of link prediction with downstream classifications. 
HGPrompt~\cite{yu2024hgprompt} proposes a dual-template design for heterogeneous graphs.
However, the aforementioned models overlook the class imbalance issue in (heterogeneous) graphs, which is a common scenario in social media for drug trafficking activities.  
\begin{algorithm}[t]
    \caption{Training Procedure of LLM-HetGDT.}\label{alg: LLM-HetGDT}
    \textbf{Input:} Heterogeneous graph $\gG = (\gV, \gE, \gX, \gA, \gR)$ with plaintext set $\gT$, encoder $f(\cdot)$ with parameters $\phi$, LLM $M(\cdot)$\\
    \# Pre-train Stage \\
    \While{\text{not converge}}{
        Feed $\gG$ into encoder $f$ to generate node embeddings $z_i^{mp}$ and $z_i^{sc}$. \\
        Compute the pretext loss $\gL_{cl}$ through Eq.~\ref{eq: nt-xent}.
        Optimize parameters $\phi$ by minimizing the loss $\gL_{cl}$.
    }
    \# Fine-tune Stage \\
    Freeze the parameters of encoder $f$. \\
   Feed $\gG$ into LLM $M$ to obtain the augmented heterogeneous graph $\gG'$ \\
   Initialize drug trafficking and node prompts, $\rmC$ and $\rmF$. \\ 
    \For{\text{each epoch}}{
       Obtain prompted node features by applying feature prompt to node attribute features in $\gG'$.  \\
       Feed the prompted node features into HGNNs $f_{\phi^*}$ to generate target node embeddings $\rmZ$. \\
       Compute the structure prompt $\rmS$ via Eq.~\ref{eq: structure} - Eq.~\ref{eq: structure prompt2}. \\
       Update target node embeddings by applying the structure prompt. \\
    Compute the total loss $\gL_{pt}$ via Eq.~\ref{eq: pt loss}. \\
    Optimize the parameters by minimizing $\gL_{pt}$.
    }   
\end{algorithm}
\begin{table*}[htbp]
    \centering
      \setlength\tabcolsep{20pt}
    \begin{tabular}{l l}
        \toprule
         \multicolumn{2}{c}{Content Feature}\\ 
         \midrule\midrule
         User (U) &  username, user ID, profile, and \# of followers.\\
         Tweet (T) & text content, \# of likes, reply, and retweet.\\
         Keyword (K) & keywords extracted from text contents.\\
         \midrule 
         \multicolumn{2}{c}{Relation}\\
         \midrule\midrule
         R1: User-follow-User & R2: User-post-Tweet  \\
         R3: User-engage-Tweet & R4: User-profile-Keyword\\
         R5: Tweet-include-Keyword & R6: Tweet-tag-Keyword\\
         \bottomrule
    \end{tabular}
        \vspace{-3mm}
     \caption{Content and relation information in heterogeneous graph dataset Twitter-HetDrug.}    \label{tab:twitter-hetdrug}
        \vspace{-4mm}
\end{table*}

\section{Data Description} \label{appendix: data description}
\subsection{HG Construction on Social Media Data}\label{sec: hg construction} 
To comprehensively describe the drug trafficking activities on social media, we construct a drug trafficking HG that integrates informative content features and complex relationships among online users. 
In this paper, we regard Twitter as an example to study online drug trafficking activities.
Specifically, we build an HG, with three types of nodes (user, tweet, and keyword) and six types of relations as well as features of each node, such that both content and relation information can be exploited simultaneously.
The details of content features and relationship information are discussed below.\\
\noindent\textbf{Content Feature.} 
To comprehensively characterize each target node (user node), unlike existing works~\cite{liu2021pick, HyGCL-DC} that merely consider users as nodes, we introduce three types of nodes, i.e., User (U), Tweet (T), and Keyword (K), into HG.
Specifically, for each node $v_i^{A}\in\gV_A$, we concatenate the text content (as listed in Table~\ref{tab:twitter-hetdrug}), denoted as $t_i^A$, and further leverage the pre-trained transformer-based language model, SentenceBert~\cite{reimers2019sentence}, to convert the concatenated text information into a fixed-length feature vector ($d=384$) 
as attribute feature $x_i^{A}$. Here, $x_i^A$ denotes the attribute feature of node $v_i^A$ belongs to node set $\gV_A$ with node type $A\in\gA$, and $\gA = \{\text{U, T, K}\}$.\\
\noindent\textbf{Relations.} To determine whether a user is a drug trafficking participant on social media, we consider not only the content-based features but also the complex relationships among users, tweets, and keywords. 
Particularly, we define six kinds of relationships (R1-R6 in Table~\ref{tab:twitter-hetdrug}): \textit{R1: user-follow-user} denotes a user is following or followed by another user; \textit{R2: user-post-tweet} represents a user posts a tweet; \textit{R3: user-engage-tweet} denotes that a user engages into a tweet, including likes, replies, retweet, and mention; \textit{R4: user-profile-keyword} represents that a user's profile contains the keyword; \textit{R5: tweet-include-keyword} denotes that a tweet content includes a keyword in plaintext; \textit{R6: tweet-tag-keyword} indicates that a tweet tags a keyword as a hashtag. 
To summarize, we build a HG $\gG = (\gV, \gE, \gX, \gA, \gR)$ by integrating content features of three kinds of entities and six types of relationships among these entities. The data statistic of the constructed HG is listed in Table~\ref{tab: data static}. 
\subsection{Data Collection on Social Media}
We crawl metadata through official Twitter API from Dec 2020 to Aug 2021. Afterward, following the existing work~\cite{HyGCL-DC, qian2021distilling}, we generate a keyword list that covers 21 drug types that may cause drug overdose or drug addiction problems to filter the tweets that contain drug-relevant information. The drug types and partial drug keywords are listed in Appendix~\ref{appendix: drug types}. Based on the keyword list, we obtain 54,680 users, with corresponding 326,826 tweets and 288 valid keywords. 
\subsection{Annotation Rules}  \label{appendix: annotation}
With the filtered metadata, six researchers spend 62 days annotating these Twitter users separately, following the annotation rules discussed below:
(i) If a user actively promotes some type of drug on Twitter or has rich connections (e.g., following, replying, liking, and retweeting) with other drug-related users in specific drug communities, he/she will be considered a drug trafficking participant. (ii) If a user appears to suffer from drug overdose or drug addiction to the specific drug, he/she will identify as a drug trafficking participant. (iii) If we can find evidence on Twitter that a user used to purchase specific drugs from others on Twitter, then he/she is considered a drug trafficking participant. For these Twitter users with disagreed labels, we conducted further discussion among annotators for cross-validation. Based on the above strategy, we can obtain the ground truth for the drug trafficking detection task.
\subsection{Drug Types and Keywords} \label{appendix: drug types}
Following previous works~\cite{qian2021distilling, HyGCL-DC}, we generate a keyword list that covers 21 drug types that may cause drug overdoes or drug addiction problems to filter the tweets that contain drug-relevant information. 
Table~\ref{tab: drug keyword} lists the drug types and partial drug keywords. 
\begin{table}[htbp]
    \vspace{-2mm}
    \setlength\tabcolsep{18pt}
    \renewcommand{\arraystretch}{0.5}
    \begin{tabular}{c|c| c}
        \toprule
         \textbf{Type} & \textbf{Category} & \textbf{Count}   \\
        \midrule\midrule
         \multirow{3}{*}{\texttt{Node}} & User & 28,839 \\
                  & Tweet  & 118,697     \\
         & Keyword & 288    \\
         \midrule
           \multirow{2}{*}{\texttt{Class}} & Participant & 3,437 \\
          & Benign & 25,402
          \\   
        \midrule
        \multirow{6}{*}{\texttt{Edge}} & R1 & 100,571 \\
        & R2 & 118,697  \\
        & R3 & 44,197 \\
        & R4 & 64,997\\
        & R5 & 260,868\\
        & R6 & 39,129\\
          \bottomrule
    \end{tabular}
    \vspace{-3mm}
     \centering    
    \caption{Statistics of the constructed HG. The underlined node type is the target node for classification.}\label{tab: data static}
    \vspace{-3mm}
\end{table}

\begin{table}[htbp]
    \setlength\tabcolsep{2pt}
    \renewcommand{\arraystretch}{0.9}
    \begin{tabular}{l|l}
        \toprule
        \textbf{Drug Type} & \textbf{Drug Name (Keywords)} \\
        \midrule  \midrule
        \texttt{Cannabis} &	Cannabis, and infused products \\
        \texttt{Opioid} &	Oxycodone, Codeine, etc. \\
        \texttt{Hallucinogen}	& LSD, MDMA, Shroom, DMT, etc. \\
        \texttt{Stimulant} & Cocaine, Amphetamine, etc. \\
        \texttt{Depressant} &	Xanax, Valium, Halcion, etc. \\
    \bottomrule
    \end{tabular}
    \vspace{-3mm}
    \centering
    \caption{Drug types and partial drug keywords.}
    \label{tab: drug keyword}
    \vspace{-3mm}
\end{table}
\section{Prompt Templates}\label{appendix: prompt template}
In this section, we provide the node prompt template in Table~\ref{tab: node prompt} and edge prompt template in Table~\ref{tab: edge prompt}.
\begin{table}[t]
\vspace{-2mm}   
    \begin{tabular}{p{0.95\linewidth}}
       \toprule
        \textbf{INPUT: }\\
        \midrule
        \textbf{Instruction:} 
        Act as an expert in social network analysis, your task is to generate a synthetic user based on the given context, including username, user ID, and user profile. The generated synthetic user information should capture the \textbf{key information} in the original user while \textbf{ensuring variability}. \\
        \textbf{Context:} 
        Username: \textbf{<Username>}; 
        User ID: \textbf{<User ID>}; 
        User Profile: \textbf{<User Profile>}.\\
       \bottomrule
    \end{tabular}  
    \vspace{-3mm}
     \centering
    \caption{The prompt template in LLM-HetGDT to generate synthetic online drug trafficking user information.} \label{tab: node prompt}
    \vspace{-4mm}
\end{table}

\begin{table}[t]
    \begin{tabular}{p{0.95\linewidth}}
       \toprule
        \textbf{INPUT: }\\
        \midrule
        \textbf{Instruction:} 
        Act as an expert in social network analysis; given the original user information, the connected neighbor information with neighbor type, and synthetic user information generated from the original user, your task is to determine which neighbors the synthetic user should connect to. \\
        \textbf{Context:} 
        Original User Information: \textbf{<Original User Information>}; 
        Neighbor Type: \textbf{<Neighbor Type>}; 
        Neighbor Information: [\textbf{<Neighbor Information 1>}, \dots ];
        Synthetic User Information: \textbf{<Synthetic User Information>}.\\
       \bottomrule
    \end{tabular}   
    \vspace{-3mm}
     \centering
    \caption{The prompt template in LLM-HetGDT to generate neighbors of synthetic online drug trafficking users.} \label{tab: edge prompt}
    \vspace{-4mm}
\end{table}

\section{Additional Mechanisms Details}\label{appendix: mechanism}
\subsection{Attention Mechanism} ~\label{appendix: attention}
The attention mechanism $\textbf{Att}(\cdot)$ in Equation~\ref{eq: prompt node} is defined as:
\begin{align}     
    \textbf{Att}(\rmX, \rmF)
     = \sum_{k=1}^K  \frac{\exp(\sigma( (\rvf_k)\top \cdot \rmX_{i, :}))}{\sum_{j=1}^K \exp(\sigma((\rvf_j)^\top \cdot \rmX_{i, :})} \cdot\rvf_k, \label{eq: att}
\end{align}
where $\sigma$ denotes the activation function, and $\rvf_k$ is the $k$-th row of the feature token $\rmF$.

\subsection{Set Transformer} ~\label{appendix: set transformer}
Here, we present set transformer in Equation~\ref{eq: prompt node}.
Particularly, the set transformer $\textbf{PMA}(\cdot)$ is defined as:
\begin{align}
    &\textbf{PMA}(\gS) = \text{LN}(\rmQ + \text{MLP}(\rmQ)), \label{eq: pma} \\ 
    &\rmQ = \text{LN}(\psi + \text{MH}(\psi, \rmY, \rmY)),
\end{align}
where $\gS$ is the set of input tokens, $\text{LN}(\cdot)$ denotes the layer normalization, $\text{MLP}(\cdot)$ is multi-layer perception, $\psi\in\sR^{d_T}$ is a learnable parameters, served as a seed (query) vector, $\rmY\in\sR^{|\gC|\times d_T}$ is the stacked embeddings of set $\gS$, 
and $\text{MH}(\cdot)$ represents the multi-head attention mechanism. 
Besides, the multi-head attention mechanism $\text{MH}(\cdot)$ is formulated as:
\begin{align}
    &\text{MH}(\psi, \rmY, \rmY) = ||_{j=1}^h \omega(\psi_j\cdot \rmK_j^\top)\rmV_j,\\
    &\rmK_j = \rmY W_j^K, \rmV_j = \rmY W_j^V,
\end{align}
where $||$ denotes the concatenation operation, $h$ is the number of heads, $\omega$ is the softmax function, and $\psi_j \in\sR^{d_T^M}$ with dimension $d_T^M = d_T / h$ .
Here $\psi = ||_{j=1}^h \psi_j$, $W_j^K\in\sR^{d_T\times d_T^M}$ and $W_j^V\in\sR^{d_T\times d_T^M}$ are the weight matrices for key and value vectors, respectively. 
\subsection{Meta Path}\label{appendix: meta path}
A meta-path $P$ is a path defined in the form of $A_1 \xrightarrow{R_1} A_2 \xrightarrow{R_2} \dots \xrightarrow{R_l} A_{l+1}$, where $A_i\in\gA$ and $R_i\in\gR$. It describes a composite relation $R = R_1 \circ R_2 \circ \dots \circ R_l$ between nodes $A_1$ and $A_{l+1}$. Here $\circ$ denotes the composition operator.
In Twitter-HetDrug, we define the following three types of meta paths:
\begin{itemize}
    \item  User-Tweet-User.
    \item  User-Tweet-Keyword-Tweet-User.
    \item  User-Keyword-User. 
\end{itemize}
\subsection{Semantic Attention Mechanisms} ~\label{appendix: semantic attention}
The semantic attention mechanism $\textbf{SM-Att}(\cdot)$ in Equation~\ref{eq: structure} is defined as:
\begin{align} 
    \textbf{SM-Att}(\gH) &=
       \sum_{\rmH_i\in\gH} \alpha_i \cdot \rmH_{i}, \text{where }\\ 
    \alpha_i &= \frac{\exp(w_i)}{\sum_{H_j\in\gH} \exp(w_j)}, \text{and } \\
    w_i &= \frac{1}{|\gH|} \sum_{\rmH_i\in \gH} q^\top \cdot \tanh(W\cdot \rmH_{i} + b), \label{eq: structure prompt2}
\end{align}
where $q\in\sR^{d}$ is the semantic-level attention vector, $W\in\sR^{d\times d}$ denotes the weight matrix, and $b\in\sR^{d}$ is the bias vector. 
The parameters $W$ and $b$
are shared across all meta-paths.  

\begin{table*}[t]
\vspace{-2mm}
 
    \setlength\tabcolsep{5pt}
    \renewcommand{\arraystretch}{0.5}
    \begin{tabular}{c|l|c|c|c|c|c|c}
    \toprule
      \multicolumn{2}{c|}{Setting} & \multicolumn{2}{c|}{Twitter-HetDrug 10\%} &  \multicolumn{2}{c}{Twitter-HetDrug 20\%} & \multicolumn{2}{|c}{Twitter-HetDrug 40\%} \\
      \midrule
      Group  & Model &  Macro-F1 & GMean &  Macro-F1 & GMean &  Macro-F1 & GMean  \\
       \midrule\midrule
       \multirow{3}{*}{\texttt{G1}} & GCN  & 55.74 \tiny{± 3.71} & 53.92 \tiny{± 2.45} & 58.38 \tiny{± 2.39}  & 56.43 \tiny{± 2.46} & 60.94 \tiny{± 2.33} & 59.72 \tiny{± 2.07} \\
       & GAT & 54.24 \tiny{± 3.53}  & 52.06 \tiny{± 2.52} & 57.93 \tiny{± 3.38} & 55.04 \tiny{± 2.74} & 60.63 \tiny{± 2.80} & 59.30 \tiny{± 2.75}  \\
       & GIN & 56.92 \tiny{± 1.14} & 54.26 \tiny{± 2.37} & 59.15 \tiny{± 1.61} & 57.92 \tiny{± 2.37} & 61.09 \tiny{± 3.74} & 60.67 \tiny{± 2.83}\\
       \midrule\midrule
          \multirow{3}{*}{\texttt{G2}} & HetGNN& 58.52 \tiny{± 3.56} & 59.21 \tiny{± 2.36} & 60.86 \tiny{± 2.71} & 61.39 \tiny{± 2.96} &  63.46 \tiny{± 3.01} & 64.05 \tiny{± 3.13} \\
          & HAN & 60.25 \tiny{± 0.79} & 59.26 \tiny{± 2.72} & 63.58 \tiny{± 0.95} & 62.07 \tiny{± 2.47} & 66.71 \tiny{± 0.92}  & 65.98 \tiny{± 3.04} \\
          & HGT  & 60.95 \tiny{± 1.61} & 59.88 \tiny{± 2.09} & 63.85 \tiny{± 1.95} & 63.70 \tiny{± 2.59}  & 67.08 \tiny{± 2.06} & 66.37 \tiny{± 3.84} \\
          \midrule\midrule
          \multirow{4}{*}{\texttt{G3}} & DMGI  & 60.53 \tiny{± 3.62} & 59.04 \tiny{± 2.65} & 63.26 \tiny{± 3.19}  & 62.94 \tiny{± 2.90}  &  65.28 \tiny{± 2.57} & 64.95 \tiny{± 3.05}  \\
           & HeCo   & 62.93 \tiny{± 2.02} & 60.85 \tiny{± 1.05} & 64.83 \tiny{± 2.53} & 63.21 \tiny{± 3.62} & 67.94 \tiny{± 2.92} & 66.52 \tiny{± 2.69} \\  
           & HGMAE   & 61.84 \tiny{± 2.61} & 60.01 \tiny{± 2.54} & 62.53 \tiny{± 3.01} & 61.73 \tiny{± 3.61} & 65.74 \tiny{± 2.46} & 64.63 \tiny{± 2.90}\\
           & MetaHG & 61.77 \tiny{± 2.45} & 60.35 \tiny{± 2.98} & 62.16 \tiny{± 3.58} & 62.50 \tiny{± 3.21} & 66.41 \tiny{± 2.07} & 65.39 \tiny{± 2.47}\\
           \midrule\midrule
          \multirow{3}{*}{\texttt{G4}} & Re-weighting & 60.72 \tiny{± 0.89} & 60.31 \tiny{± 0.94} & 62.38 \tiny{± 0.52} & 63.07 \tiny{± 0.58} &  65.05 \tiny{± 0.59} & 65.91 \tiny{± 0.46} \\
                              & Over-sampling &  63.79 \tiny{± 0.73} & 62.27 \tiny{± 0.75} & 65.10 \tiny{± 0.32} &  63.08 \tiny{± 0.38} & 68.35 \tiny{± 0.40} & 67.39 \tiny{± 0.85} \\
                              & SMOTE  & 64.25 \tiny{± 0.72} & 62.17 \tiny{± 0.86} & 66.94 \tiny{± 0.82} & 63.62 \tiny{± 0.77}  & 69.24 \tiny{± 0.84} & 66.77 \tiny{± 0.93} \\
        \midrule\midrule
          \multirow{3}{*}{\texttt{G5}} & GraphENS & 65.37 \tiny{± 1.72}  & 63.80 \tiny{± 3.49} & 66.93 \tiny{± 0.94}  &  64.06 \tiny{± 3.19} & 68.37 \tiny{± 1.37}  &  70.82 \tiny{± 2.29}\\
                              & GraphSMOTE & 65.82 \tiny{± 3.75} & 64.04 \tiny{± 2.39} & \underline{68.07 \tiny{± 3.89}} & 65.83 \tiny{± 2.63} & 70.32 \tiny{± 2.46} & 69.25 \tiny{± 2.09} \\
                              & PC-GNN  & 65.98 \tiny{± 2.83}  & 63.84 \tiny{± 3.48} & 67.52 \tiny{± 2.95}  & 64.71 \tiny{± 3.04}  &  \underline{70.59 \tiny{± 2.48}}  & \underline{71.94 \tiny{± 3.74}} \\
         \midrule\midrule
          \multirow{4}{*}{\texttt{G6}} & GPPT & 62.84 \tiny{± 4.72} & 60.41 \tiny{± 2.40} & 64.97 \tiny{± 3.47} & 63.28 \tiny{± 2.53}  & 68.04 \tiny{± 2.46}  & 67.42 \tiny{± 2.73} \\
                              & Self-Pro & 63.37 \tiny{± 3.52}  &  61.05 \tiny{± 3.61} & 65.23 \tiny{± 3.14}  & 63.94 \tiny{± 3.51} & 68.33 \tiny{± 2.73} & 67.16 \tiny{± 1.93} \\
                              & GraphPrompt & 64.37 \tiny{± 1.37} & 62.38 \tiny{± 3.52} & 65.89 \tiny{± 2.85}  & 64.53 \tiny{± 1.92}  & 68.62 \tiny{± 2.46}  & 68.37 \tiny{± 2.18} \\
                             & HGPrompt & \underline{66.85 \tiny{± 2.95}} & \underline{64.37 \tiny{± 2.53}} & 67.97 \tiny{± 3.52} & \underline{66.08 \tiny{± 2.08}}  & 69.72 \tiny{± 3.92} & 69.42 \tiny{± 3.24} \\
         \midrule\midrule
         \multirow{3}{*}{\texttt{G7}} & LLama 70B & 65.07 \tiny{± 0.00} & 63.84 \tiny{± 0.00} & 64.90 \tiny{± 0.00} & 63.77 \tiny{± 0.00} & 75.37 \tiny{± 0.00} & 64.02 \tiny{± 0.00}         \\ 
         & Mistral 24B & 64.83 \tiny{± 0.00} & 62.87 \tiny{± 0.00} & 64.90 \tiny{± 0.00} & 62.57 \tiny{± 0.00} & 64.73 \tiny{± 0.00} & 62.36 \tiny{± 0.00} \\ 
         & DeepSeek-R1 70B & 65.74 \tiny{± 0.00} & 63.92 \tiny{± 0.00} & 65.36 \tiny{± 0.00} & 63.98 \tiny{± 0.00} & 65.18 \tiny{± 0.00} & 63.65 \tiny{± 0.00} \\ 
         \midrule\midrule
         \rowcolor{Gray}   \texttt{Ours} & LLM-HetGDT  & \textbf{70.52 \tiny{± 1.58}}  & \textbf{71.69 \tiny{± 2.50}} & \textbf{71.36 \tiny{± 1.72}} & \textbf{73.87 \tiny{± 1.45}} &  \textbf{74.06 \tiny{± 2.74}} &  \textbf{76.24 \tiny{± 2.85}}\\
           \bottomrule
         
    \end{tabular}
    \vspace{-3mm}
       \centering \caption{Performance (mean ± std) comparison among various methods on Twitter-HetDrug for drug trafficking detection. Bolded numbers indicate the best performance, and underline numbers represent runner-up performance.} \label{tab: complete performance}
    \vspace{-3mm}
\end{table*}

\begin{table}[htbp]
    \centering
    \setlength\tabcolsep{5pt}
    \renewcommand{\arraystretch}{0.8}
    \begin{tabular}{l|c}
    \toprule
        \textbf{Hyper-parameters} &  \textbf{Values} \\
        \midrule\midrule
         \texttt{Hidden Dimension} & \{128, 256, 512, 1024\}\\
         \texttt{Number of Layers} & \{1, 2, 3\} \\
            \texttt{Normalize}  & \{none, batch, layer\} \\ 
            \texttt{Learning Rate}  & \{1e-2, 1e-3, 1e-4\} \\
            \texttt{Weight Decay} & \{0.0, 1e-5, 1e-4, 1e-3\} \\ 
            \texttt{Dropout} & \{0.0, 0.1, 0.5, 0.8\} \\ 
            \texttt{Attention Head}* & \{1, 2, 4, 8\}\\
            \bottomrule
    \end{tabular}
    \vspace{-3mm}
    \caption{The hyper-parameters we searched. Star hyper-parameters (*) are not applicable to every models.}\label{tab: hyper-parameter}
    \label{tab:my_label}
\end{table}
\begin{table*}[htbp]
    \centering
    \setlength\tabcolsep{4pt}
    \renewcommand{\arraystretch}{0.8}
    \begin{tabular}{c|c|c|c|c|c|c|c|c|c|c|c}
    \toprule
        Ratio & Dim. & Layers & Normalize & LR. & Decay & Dropout & Att. Head & \# basic vec. & $\delta$ & $\lambda$ \\
        \midrule\midrule
        10\% & 256 & 3 & layer & 1e-3 & 1e-4 & 0.1 & 4 & 10 & 5e-2 & 1e-3\\ 
        20\% & 512 & 2 & layer & 1e-3 & 1e-5 & 0.5 & 1 & 15 & 5e-2 & 1e-3\\ 
        40\% & 512 & 3 & layer & 1e-4 & 1e-5 & 0.5 & 1 & 10 & 5e-2 & 1e-3\\ 
        \bottomrule
    \end{tabular}
    \vspace{-3mm}
    \caption{The hyper-parameters for our LLM-HetGDT in basic training setting.}\label{tab: best hp}
\end{table*}
\section{Baseline Settings}\label{appendix: baseline}
We compare LLM-HetGDT with seven groups of baseline methods, including graph representation learning methods (G1), heterogeneous graph representation learning models (G2), heterogeneous graph self-supervised learning methods (G3), generic methods against class imbalance data (G4), graph models against the class imbalance issue (G5), graph prompt learning models (G6), and LLMs-based methods (G7). 
Next, we would like to introduce these baseline methods in detail.\\
\noindent\textbf{G1 Graph Representation Learning Methods:} 
We implement Graph Convolution Network (GCN)~\cite{GCN}, Graph Attention Network (GAT)~\cite{velivckovic2017graph}, and Graph Isomorphism Network (GIN)~\cite{GIN}. We employ a two-layer MLP as a classifier at the end of these models for the node classification tasks.\\
\noindent\textbf{G2 Heterogeneous Graph Representation Learning Models:}
We choose three state-of-the-art (SOTA) heterogeneous graph representation learning methods, including Heterogeneous Graph Neural Network (HetGNN)~\cite{zhang2019heterogeneous}, Heterogeneous Graph Attention Network (HAN)~\cite{wang2019heterogeneous}, and Heterogeneous Graph Transformer (HGT)~\cite{hu2020heterogeneous}.
Specifically, HetGNN utilizes random walks to sample node neighbors and further fuse their attribute features using LSTMs to generate node embeddings. 
HAN converts multiple metapath-based homogeneous graphs and leverages hierarchical attention to encode node-level and semantic-level structures to obtain node embeddings.
HGT adopts a transformer-based architecture for web-scale heterogeneous graphs. 
Similarly, we utilize two-layer MLP as a classifier over the node embeddings for the node classification tasks.\\
\noindent\textbf{G3 Heterogeneous Graph Self-supervised Learning Methods:} 
We also implementfour self-supervised methods for heterogeneous graph representation learning, i.e., Deep Graph Infomax for Attribute Multiplex Network (DMGI)~\cite{park2020unsupervised}, Heterogeneous Graph Neural Network with Co-contrastive Learning (HeCo)~\cite{wang2021self}, Heterogeneous Graph Masked Autoencoder (HGMAE)~\cite{tian2023heterogeneous}, and MetaHG~\cite{qian2021distilling}.
Particularly, DMGI introduces a consensus regularization to minimize the disagreements among the relation-type specific node embeddings and a universal discriminator that discriminates true samples regardless of the relation types. 
HeCo adopts cross-view contrastive learning on heterogeneous graphs, i.e., meta-path and network schema view encoders.
HGMAE leverages two masking strategies toward meta-path adjacency matrix and attributes features with three training techniques, i.e., meta-path edge reconstruction, target attribute restoration, and positional feature prediction, to obtain target node embeddings.
MetaHG is a meta knowledge distillation framework over heterogeneous graphs for drug trafficking detection.
\\
\noindent\textbf{G4: Generic Methods Against Class Imbalance Data: } 
We compare LLM-HetGDT with three classic generic methods for class imbalance data. 
Re-weighting~\cite{cui2019class} scales the training loss with respect to the weight of classes.
Over-sampling~\cite{kang2019decoupling} duplicates the samples from minority classes to balance the training data distribution.
SMOTE~\cite{chawla2002smote} is an over-sampling method that generates synthetic samples from minority classes by interpolating between existing samples.\\
\noindent\textbf{G5 Graph Models Against the Class Imbalance Issue:} 
We select three SOTA graph-based methods for class imbalance issues, including GraphENS~\cite{park2022graphens}, GraphSMOTE~\cite{zhao2021graphsmote}, and PC-GNN~\cite{liu2021pick}. 
GraphENS synthesizes the whole ego network for the minor class (minor node and its one-hop neighbors) by combining two different ego networks based on their similarity and further introduces a saliency-based node mixing method to exploit the abundant class-generic attributes of other nodes while blocking the injection of class-specific features.
GraphSMOTE extends SMOTE to the graph data by designing a re-sampling module and an edge predictor to determine the connectivity between synthetic nodes and other nodes in graphs. 
PC-GNN devises a pick-and-choose strategy to handle the class imbalance issue in multi-relation graphs. \\
\noindent\textbf{G6 Graph Prompt Learning Models:}
We also compare our proposed method with four (heterogeneous) graph prompt learning baseline methods, i.e., GPPT~\cite{sun2022gppt}, Self-Pro~\cite{gong2024selfpro}, GraphPrompt~\cite{liu2023graphprompt}, and HGPrompt~\cite{yu2024hgprompt}. 
GPPT and GraphPrompt introduce prompt templates to align the pretext task of linked prediction with downstream classification. 
Self-Pro leverages asymmetric graph contrastive learning as a pretext to address heterophily and align the objective of pretext and downstream tasks. It also proposes to reuse the component from pre-training as an adapter and introduce self-prompts based on the graph for task adaptations. 
HGPrompt designs a dual prompt template for both homogeneous and heterogeneous graphs. 
Mention that for all baseline methods that are designed for homogeneous graphs, following existing works~\cite{wang2019heterogeneous, park2020unsupervised, tian2023heterogeneous}, we directly use all meta-paths with target node attribute features as homogeneous graphs for downstream tasks.
Additionally, we strictly follow the source code to reproduce the experiment result.\\
\noindent\textbf{G7 LLMs-based Methods:}
As there are only one LLMs-based method, i.e., KG-GPT~\cite{hu2024knowledge}, for drug trafficking detection, we employ the prompt in KG-GPT with different backbone LLMs as baseline methods. 
Specifically, we choose three state-of-the-art LLMs, including LLama-3.1-70B-Instruct~\cite{llama3.1}, Mistral-Small-24B-Instruct~\cite{Ministral8BInstruct2410}, and DeepSeek-R1-Distill-Llama-70B~\cite{deepseekai2025deepseekr1incentivizingreasoningcapability}.

\section{Additional Experimental Settings}\label{appendix: experimental settings}
We follow the existing works~\cite{wang2024can, qian2021distilling} to grid search the optimal hyper-parameter of each model, and report the best performance. 
The hyper-parameters we search for are listed in Table~\ref{tab: hyper-parameter}.
The optimal hyper-parameters of our model are provided in Table~\ref{tab: best hp}.

\section{Deployment Evaluation Settings} \label{appendix: deployment}
To further validate the applicability of our model, we employ LLM-HetGDT to detect drug trafficking activities in a live manner. 
Specifically,  we first identify drug sellers in our dataset, i.e., Twitter-HetDrug, and further leverage these anchor users to expand their social circles over Twitter. 
Then we follow the same procedure in Appendix~\ref{appendix: data description} to construct the HG for the expanded social circles and annotate the labels for the users. 
The constructed HG contains 416 users, including 61 drug trafficking participants and 355 benign users, 3,209 tweets, and 84 keywords. 
Afterward, we employ baseline methods and our proposed method to detect drug trafficking participants over the constructed HG. 
Mention that, we randomly choose three drug trafficking participants and seven benign users as training set, and the rest are for testing set.
To ensure a fair comparison between supervised methods and LLM-based approaches, we include the three drug trafficking participants from the training set as few-shot examples within the LLM prompt during inference.

\section{Additional Experiment Comparison} \label{appendix: complete experiments}
Due to the space limit in main paper, we provide additional experiment results in this section. 
Table~\ref{tab: complete performance} lists the performance of all baseline methods and our proposed method over Twitter-HetDrug.

\begin{table}[htbp]
    \vspace{-3mm}
        \centering
    \setlength\tabcolsep{4pt}
    \renewcommand{\arraystretch}{0.6}
    \begin{tabular}{c | c | c | c | c}
           \toprule 
           \# of basis vectors & 1 &  5 & 10 & 15 \\ 
        \midrule\midrule
        Twitter 10\% & 66.83  & \underline{69.78} & \textbf{70.52} & 69.32 \\ 
        \midrule
           Twitter 20\% & 68.39 & 70.69  &  \underline{71.04} &  \textbf{71.36} \\ 
        \midrule 
           Twitter 40\% & 71.24 & \underline{73.72} & \textbf{74.06} & 73.58 \\ 
        \bottomrule
         \toprule
        Hidden dimension &  128 & 256 & 512 & 1,024\\
                \midrule\midrule
        Twitter 10\% &  69.74 & \textbf{70.52} &  \underline{70.28} & 69.84 \\ 
        \midrule
        Twitter 20\% & 70.36 & \underline{71.09}  &   \textbf{71.36}  & 70.48\\ 
        \midrule 
        Twitter 40\% & 73.85 & \underline{73.97} &  \textbf{74.06} & 73.89 \\ 
        \bottomrule
    \end{tabular}
    \vspace{-3mm}
     \caption{Sensitivity test (Macro-F1) on the hyper-parameters: \# of basis vectors and hidden dimensions. }
    \label{tab: hyper-parameter sensitivity 1}
    \vspace{-3mm}
\end{table}

\section{Hyper-parameter Sensitivity Analysis}
We conduct hyper-parameter sensitivity experiments among Twitter-HetDrug. Specifically, we control the number of basis vectors in node prompts, i.e., $K$, and evaluate the performance in Macro-F1 of LLM-HetGDT. The performance is shown in Table~\ref{tab: hyper-parameter sensitivity 1}. 
The bolded numbers indicate the best performance, and the underlined numbers represent the runner-up performance.
According to Table~\ref{tab: hyper-parameter sensitivity 1}, we find that even a relatively small number ($K=5$) of basic vectors can achieve good performance in drug trafficking detection.
The best performance is achieved at $K=10$ for Twitter-HetDrug with train split $10\%$ and $40\%$, and shows satisfactory performance for Twitter-HetDrug with train split $20\%$. 
This indicates the effectiveness of our model, as it achieves excellent performance with a relatively small number of basis vectors. 
We also perform sensitivity analysis to show the impact of hidden dimensions. 
According to Table~\ref{tab: hyper-parameter sensitivity 1}, we find that the optimal hidden dimension can be different across Twitter-HetDrug with different training splits. 
For example, the optimal hidden dimension for the 10\% training split is 256, while 512 works best for the 20\% and 40\% splits. However, increasing the hidden dimension to 1024 leads to reduced model performance, possibly because a wider hidden layer may prevent the model from focusing on the most relevant information.
Hence, using standard hidden dimensions, i.e., 256 or 512, is sufficient for LLM-HetGDT to capture the complex information, which again shows the superiority of our model LLM-HetGDT. 
\section{Additional Case Study} \label{appendix: case study}
We further conduct a case study for edge generation. 
Specifically, after obtaining synthetic users, we employ LLM to generate edges between synthetic users and the neighbors of corresponding labeled users.
Table~\ref{tab: case study edge generation} illustrates the edges generated for synthetic users in Section~\ref{sec: case study}.
According to Table~\ref{tab: case study edge generation}, 
the original user has four neighboring tweets, consisting of one daily update and three advertising posts.
In the response, the LLM accurately identifies the last three advertising posts as the neighbors for the synthetic user while disregarding the first. 
Moreover, it offers a logical explanation for each selection. 
For instance, the third tweet, "Same day shipping in ***," lacks any drug-related content, which might be ignored by edge generation methods at the attribute feature level~\cite{zhao2021graphsmote, liu2021pick}. 
However, the LLM can accurately recognize this tweet as a promotion post for the same-day shipping service, effectively capturing the underlying behavior of the synthetic user.
\begin{table*}[htbp]
   
    \renewcommand{\arraystretch}{0.9}
    \begin{tabular}{p{0.95\linewidth}}
        \toprule
        \textbf{INPUT: }\\
        \midrule
        \textbf{Instruction:}\\
        Act as an expert in social network analysis; given the original user information, the connected neighbor information with neighbor type, and synthetic user information generated from the original user, your task is to determine which neighbors the synthetic user should connect to. \\ 
        \textcolor{blue}{Please provide the ration behind your selection.}\\
        \textbf{Context:} \\
        \textit{Original User Information:}
        Username: yas***exb; User ID: @dru***cy; 
        User Profile: Online dispensary, premium lab-tested psychedelic products, must be 18+, same day shipping in ***. DM me for menu. \#LSD \#420 \#acid. \\ 
        \textit{Neighbor Type:}
        Tweet. \\  
        \textit{Neighbor Information:} \\ 
        1. Text Content: Vibe for everyday. How was your day?\\ 
        2. Text Content: DM me for menu. \#Meth\#LSD\#420\#AcidTrip.\\ 
        3. Text Content: Same day shipping in ***.\\
        4. Text Content: !!!Giveaway!!! I will be giving away ONE 12.30OZ BAG OF METH. To enter, - Follow me, Retweet, and Reply 'Meth' under this post. END AUG ***. \\ 
        \textit{Synthetic User Information:} \\ 
        Username: Psy***ete; User ID: @ss***ete; 
        User Profile: Your one-stop shop for lab-tested psychedelic products. Get ready to trip with the selection of premium psychedelic products. \#lsdtrip \#psychedelics \#acid \#weed \#420 \#mushrooms.\\
        \midrule
        \textbf{OUTPUT: }\\
        \midrule
        Selected Neighbors: [2, 3, 4].\\
        Reasons:\\ 
        Post 2 contains the hashtags \#Meth, \#LSD, \#420, and \#AcidTrip, and \textcolor{blue}{these drug names match the products in the profile.}\\ 
        Post 3 \textcolor{blue}{promotes the same-day shipping service in location ***}, which is most likely a post by the synthetic user.\\
        Post 4 is a giveaway post about meth. This post mainly focuses on \textcolor{blue}{giveaway and promotion} the products to attract potential buyers, which is highly relevant to the synthetic user's profile.\\
       \bottomrule
    \end{tabular}
    \vspace{-3mm}
     \centering 
    \caption{Example of neighbors selected by LLM.} \label{tab: case study edge generation}  
    \vspace{-3mm}
\end{table*}

\newpage
\appendix

\end{document}

%% file: math_commands.tex
\usepackage{amsmath,amsfonts,bm}

\def\eqref#1{equation~\ref{#1}}

\def\1{\bm{1}}

\def\rvf{{\mathbf{f}}}

\def\rvn{{\mathbf{n}}}

\def\rvp{{\mathbf{p}}}

\def\rmC{{\mathbf{C}}}

\def\rmF{{\mathbf{F}}}

\def\rmH{{\mathbf{H}}}
\def\rmI{{\mathbf{I}}}

\def\rmK{{\mathbf{K}}}

\def\rmQ{{\mathbf{Q}}}

\def\rmS{{\mathbf{S}}}

\def\rmV{{\mathbf{V}}}

\def\rmX{{\mathbf{X}}}
\def\rmY{{\mathbf{Y}}}
\def\rmZ{{\mathbf{Z}}}

\DeclareMathAlphabet{\mathsfit}{\encodingdefault}{\sfdefault}{m}{sl}
\SetMathAlphabet{\mathsfit}{bold}{\encodingdefault}{\sfdefault}{bx}{n}

\def\gA{{\mathcal{A}}}

\def\gC{{\mathcal{C}}}

\def\gE{{\mathcal{E}}}

\def\gG{{\mathcal{G}}}
\def\gH{{\mathcal{H}}}

\def\gL{{\mathcal{L}}}

\def\gN{{\mathcal{N}}}

\def\gP{{\mathcal{P}}}

\def\gR{{\mathcal{R}}}
\def\gS{{\mathcal{S}}}
\def\gT{{\mathcal{T}}}

\def\gV{{\mathcal{V}}}

\def\gX{{\mathcal{X}}}

\def\sR{{\mathbb{R}}}

